\documentclass[10pt,conference,a4paper]{IEEEtran}
\IEEEoverridecommandlockouts
\usepackage{cite}
\usepackage{amsmath,amssymb,amsfonts}
\usepackage{algorithmic}
\usepackage{graphicx}
\usepackage{textcomp}
\usepackage{xcolor}
\usepackage{booktabs}
\def\BibTeX{{\rm B\kern-.05em{\sc i\kern-.025em b}\kern-.08em
    T\kern-.1667em\lower.7ex\hbox{E}\kern-.125emX}}
\makeatletter
\newcommand{\linebreakand}{%
  \end{@IEEEauthorhalign}
  \hfill\mbox{}\par
  \mbox{}\hfill\begin{@IEEEauthorhalign}
}
\makeatother

\begin{document}

\title{An Edge AI System Based on FPGA Platform for Railway Fault Detection}

\author{
\IEEEauthorblockN{Jiale Li}
\IEEEauthorblockA{\textit{School of Computer Science} \\
\textit{The University of Auckland}\\
Auckland, New Zealand \\
jli990@aucklanduni.ac.nz}
\and
\IEEEauthorblockN{Yulin Fu}
\IEEEauthorblockA{\textit{School of Computer Science} \\
\textit{The University of Auckland}\\
Auckland, New Zealand \\
yxue732@aucklanduni.ac.nz}
\and
\IEEEauthorblockN{Dongwei Yan}
\IEEEauthorblockA{\textit{School of Computer Science} \\
\textit{The University of Auckland}\\
Auckland, New Zealand \\
dyan232@aucklanduni.ac.nz}
\linebreakand 
\IEEEauthorblockN{Sean Longyu Ma}
\IEEEauthorblockA{\textit{School of Computer Science} \\
\textit{The University of Auckland}\\
Auckland, New Zealand \\
sean.ma@auckland.ac.nz}
\and
\IEEEauthorblockN{Chiu-Wing Sham}
\IEEEauthorblockA{\textit{School of Computer Science} \\
\textit{The University of Auckland}\\
Auckland, New Zealand \\
b.sham@auckland.ac.nz}
}

\maketitle

\begin{abstract}
As the demands for railway transportation safety increase, traditional methods of rail track inspection no longer meet the needs of modern railway systems. To address the issues of automation and efficiency in rail fault detection, this study introduces a railway inspection system based on Field Programmable Gate Array (FPGA). This edge AI system collects track images via cameras and uses Convolutional Neural Networks (CNN) to perform real-time detection of track defects and automatically reports fault information. The innovation of this system lies in its high level of automation and detection efficiency. The  neural network approach employed by this system achieves a detection accuracy of 88.9\%, significantly enhancing the reliability and efficiency of detection. Experimental results demonstrate that this FPGA-based system is 1.39× and 4.67× better in energy efficiency than peer implementation on the GPU and CPU platform, respectively.
\end{abstract}


\section{Introduction}
The safety and reliability of railway infrastructure are paramount given the global dependency on rail transportation for cargo and passenger movement~\cite{yuan2019deep, Li2015ImprovementsIA, Li2012AVD, Li2012AR}. Traditional rail inspection methods are increasingly unable to meet the high standards required by modern rail systems. The limitations of manual inspection-such as its time-consuming nature and susceptibility to human error-necessitate more sophisticated, automated solutions~\cite{zheng2021defect, Papaelias2008ARO, Bosso2012ACS}. 

Recent advancements in railway track fault detection leverage the capabilities of deep learning and neural networks~\cite{Yao2022RAPQRA, Liu2022PDQuantPQ, Yue2024UrbanAS, Liu2024ViTLOBEV, Zhao2023GSNHVNETAL} to enhance the accuracy and efficiency of inspections. Techniques like CNNs have been proven to substantially improve fault detection performance by enabling real-time processing and analysis of complex image data captured from rail tracks~\cite{zhang2021mrsdi,xu2020rail}. These methods have improved the detection rates and significantly reduced the need for human intervention, thus minimizing the risk of errors and oversights during inspections. Furthermore, the integration of FPGA technology with AI-based systems has opened new avenues for developing real-time, efficient, and highly reliable rail inspection systems~\cite{Lo2020EnergyEF}. FPGA platforms offer the necessary speed and flexibility to process large volumes of image data, making them ideal for applications requiring real-time analytics and decision-making~\cite{ma2019real}.

This study builds on these technological advancements and proposes a novel FPGA-based rail inspection system. This system utilizes a combination of cameras and lightweight CNN models to detect and classify track surface defects with high accuracy. The application of FPGA primarily aims to ensure energy-efficient processing while maintaining high operational efficiency~\cite{Brejza2017AHF, Zarubica2007MultiGbpsFL, Zhong2023JointSC, Ma2019SoCFPGABasedIO}, thereby significantly enhancing the system’s capability to handle data on the fly and respond swiftly to detected anomalies in rail inspection. The main contributions of this study are as follows:
\begin{itemize}
    \item An edge AI system based on an FPGA platform for railway fault detection is proposed to replace tedious and inefficient manual inspection.
    \item We designed a lightweight neural network for detecting rail faults, which achieved an accuracy of 88.9\%.
    \item According to our experiment, our work can achieve an energy efficiency of 3.41 GOPS/W, which is 1.39× and 4.67× more efficient than modern GPU and CPU, respectively.
\end{itemize}

\section{SYSTEM ARCHITECTURE}
\begin{figure}[htbp]
\centerline{\includegraphics[width=0.5\textwidth]{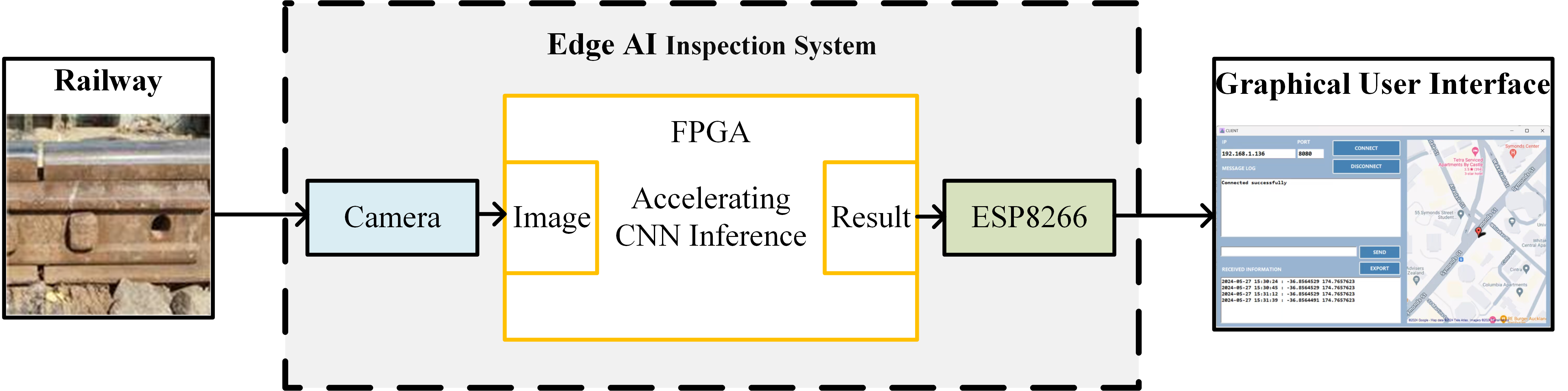}}
\caption{The edge AI system for railway fault detection}
\label{fig1}
\end{figure}
The architecture of the edge AI system we propose, as depicted in Figure 1, consists of four main components: a camera, an FPGA platform, an ESP8266 module, and a computer equipped with a graphical user interface. The system operates by capturing track images via the camera. These images are then processed on the FPGA platform, which accelerates CNN inference for efficient defect detection. Detected results are transmitted wirelessly using the ESP8266 module and displayed on the computer's graphical interface for user interaction.
\section{Methodology}
\subsection{Image Acquisition}
The images in the dataset were collected from two sources: the Railway Track Fault Detection dataset and the Surface Defect Detection dataset. The combined dataset was then split into training, testing, and validation sets, with 88\% (587 images) used for training, 6\% (40 images) for testing, and 6\% (40 images) for validation. The camera captures 128x128 pixel colour images, with each pixel value normalized from the original range of [0,255] to [0,1]. Figure 2 displays sample input images to the model and their corresponding confidence levels. The confidence levels indicate how confident the model is in its classification of each image. A higher confidence level suggests a more accurate classification by the model. In addition, during the training phase, image augmentation techniques such as blurring, noise addition, random flipping, and rotation are applied to enhance variability within the dataset.
\begin{figure}[htbp]
\centerline{\includegraphics[width=0.5\textwidth]{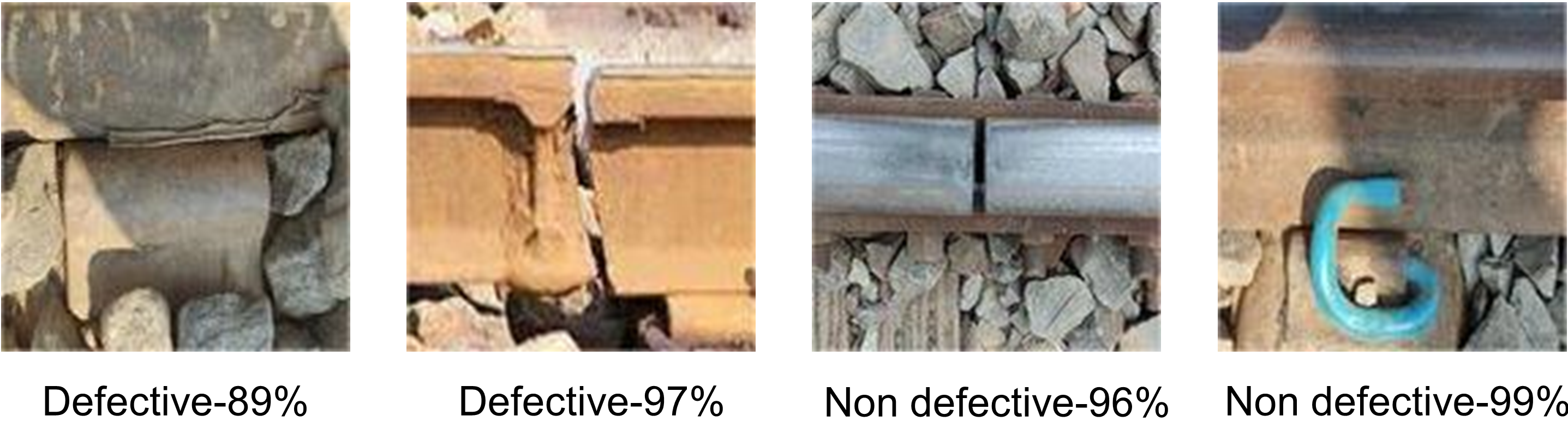}}
\caption{Sample input images to the model}
\label{fig2}
\end{figure}
\subsection{CNN Model Design}
As shown in Figure 3, this study employs a lightweight neural network to recognize patterns in rails and fasteners. Convolution layers are utilized to extract features from objects while pooling layers help reduce the dimensions of feature maps. Normalization layers and shortcut connections are incorporated to enhance the stability and efficiency of network training. A global average pooling layer is used to transform two-dimensional feature maps into one-dimensional feature vectors. Fully connected layers classify the images into distinct categories. The softmax layer converts input values into a probability distribution, and the ReLU activation function is used throughout the model to introduce non-linearity and facilitate efficient training.
\begin{figure}[htbp]
\centerline{\includegraphics[width=0.5\textwidth]{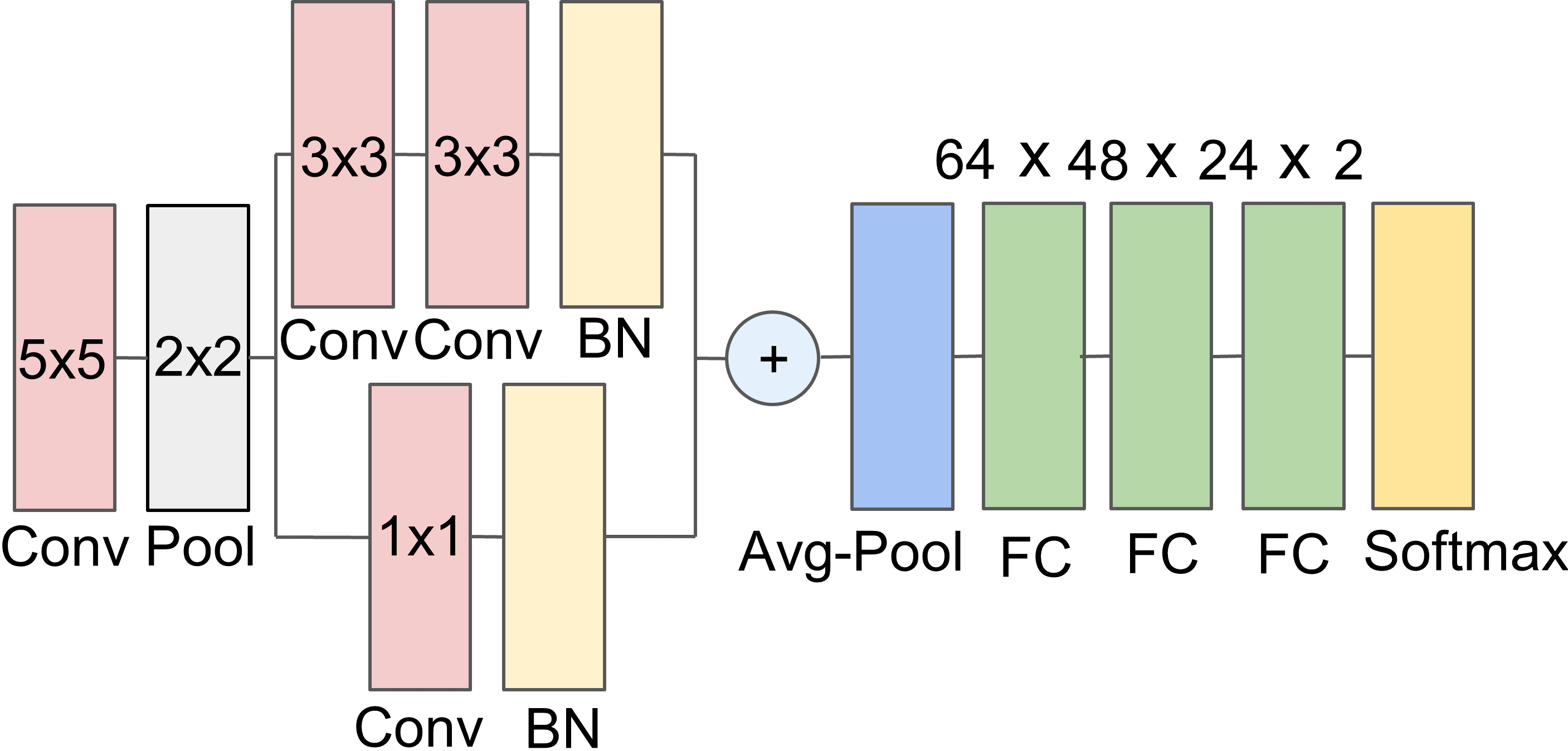}}
\caption{The lightweight neural network architecture}
\label{fig3}
\end{figure}

\subsection{Hardware Design}
The neural network inference is implemented on the FPGA side, while other tasks are handled by other modules. The FPGA's Programmable Logic (PL) part plays a pivotal role in enabling the concurrent handling of multiple operations, significantly accelerating the matrix and vector calculations critical for neural network inference. This design uses loop tiling, layered quantization, and operator fusion to optimize processing speed. In the system's architecture, the Processing System (PS) part of the FPGA is responsible for interfacing with the camera to capture images. Upon detecting any damage on the railway tracks, the PS issues a command to the ESP8266, which then transmits the location and time data to the graphical user interface.

\section{Experimental Results}
The image recognition system is implemented on the Xilinx FPGA MPSoC ZCU104. This board uses the Zynq UltraScale+ XCZU7EV-2FFVC1156 MPSoC, which contains an embedded ARM CPU. ZCU104 is supported by PYNQ, which can import and invoke the accelerator as an overlay in a Python environment. This study uses Vivado 2022.2 for FPGA block design and synthesis, Vitis HLS 2022.2 for synthesizing C/C++ functions into register transfer level (RTL) code, and PYNQ for interaction between the PS and PL. A mixed encoding method utilizing both 12-bit and 22-bit fixed-point numbers was employed, achieving an accuracy of 88.9\%, which is equivalent to that of 32-bit floating-point representations. Due to the mixed layered quantization, there is a significant reduction in hardware resource costs while maintaining nearly unchanged accuracy. The hardware resource utilization of FPGA is listed in Table I for the implementation.
\begin{table}[ht]\centering
\caption{HARDWARE RESOURCE UTILIZATION ON THE FPGA}
\label{tab:parameters}
\resizebox{\columnwidth}{!}{
\begin{tabular}{@{}lcccc@{}}
\toprule
Resource& LUT  & FF  & BRAM& DSP\\ \midrule
Value(Utilization\%)& 211388(92\%)     &  63958(14\%)    &   230.5(74\%)       & 538(31\%) \\ \bottomrule
\end{tabular}}
\end{table}

In Table II, we compare our designs with desktop GPU and CPU using the same network. Due to the camera capturing images at a rate of 60 fps, a latency below 17 ms is sufficient to meet the requirements for real-time processing. All three platforms perform well in accelerating neural networks. However, the advantage of FPGA in terms of energy consumption is clearly demonstrated. The running power of GPU and CPU are 50.6 W and 61.1 W while that of FPGA is only 6.9 W. So our design can achieve up to 1.39× better energy efficiency than GPU and 4.67× better than CPU.

\begin{table}[ht]\centering
\caption{PERFORMANCE COMPARISON OF FPGA, GPU, AND CPU}
\label{tab:parameter}
\resizebox{\columnwidth}{!}{
\begin{tabular}{@{}lccc@{}}
\toprule
            & FPGA & GPU  & CPU\\ \midrule
Platform     & ZCU104    &  RTX 4080      & i7 14700KF \\
Release Date (Year) & 2018     &   2022      & 2023 \\
Technology (nm)       & 16      &   5      & 10  \\
Frequency (MHz)           & 110      &   2205       & 5500\\
Time (ms) & 2.1     &   0.4      & 1.1 \\
Power (W)      & \textbf{6.9}       &   50.6      & 61.1  \\
Energy  Efficiency (GOPS/W)           &  \textbf{3.41}     &  2.44       & 0.73
\\ \bottomrule
\end{tabular}}
\end{table}

\section{Conclusions}
This study presented a novel FPGA-based edge AI system designed for the efficient and reliable detection of faults in railway infrastructure. The system integrates a lightweight neural network, achieving a commendable accuracy of 88.9\%. Notably, the system's energy efficiency significantly surpasses similar implementations on GPU and CPU platforms, demonstrating improvements of 1.39× and 4.67×, respectively. These results underscore the benefits of using FPGA platforms for real-time, energy-efficient railway track inspection, especially in edge scenarios where power consumption and processing speed are critical.

\bibliographystyle{IEEEtran}
\bibliography{reference}

\end{document}